\documentclass{article} 
\usepackage{iclr2024_conference,times}

\usepackage{amsmath,amsfonts,bm}

\def\eqref#1{equation~\ref{#1}}

\def\1{\bm{1}}

\DeclareMathAlphabet{\mathsfit}{\encodingdefault}{\sfdefault}{m}{sl}
\SetMathAlphabet{\mathsfit}{bold}{\encodingdefault}{\sfdefault}{bx}{n}

\usepackage{hyperref}
\usepackage{url}
\usepackage{graphicx}

\title{Loss-Free Machine Unlearning}

\author{Jack Foster$^{*1,2}$, Stefan Schoepf$^{*1,2}$ \& Alexandra Brintrup$^{1,2}$ \\
$^1$Department of Engineering, University of Cambridge\\ $^2$The Alan Turing Institute \\
*Indicates equal contribution \\
\texttt{\{jwf40,ss2823,ab702\}@cam.ac.uk} \\
}

\iclrfinalcopy 
\begin{document}

\maketitle

\begin{abstract}
We present a machine unlearning approach that is both retraining- and label-free. Most existing machine unlearning approaches require a model to be fine-tuned to remove information while preserving performance. This is computationally expensive and necessitates the storage of the whole dataset for the lifetime of the model. Retraining-free approaches often utilise Fisher information, which is derived from the loss and requires labelled data which may not be available. Thus, we present an extension to the Selective Synaptic Dampening algorithm, substituting the diagonal of the Fisher information matrix for the gradient of the $l_2$ norm of the model output to approximate sensitivity. We evaluate our method in a range of experiments using ResNet18 and Vision Transformer. Results show our label-free method is competitive with existing state-of-the-art approaches.
\end{abstract}

\section{Introduction \& Related Work}
With increasing regulations around machine learning, the ability for a trained model to forget private or harmful information is essential. This task is typically formulated by splitting some training dataset $\mathcal{D}$ into a subset of data that needs to be forgotten $\mathcal{D}_f$, and the retained data $\mathcal{D}_{r} = \mathcal{D} \setminus \mathcal{D}_{f}$. The objective is to forget $\mathcal{D}_f$ from the model, while simultaneously preserving model performance on $\mathcal{D}_r$ \citep{cao2015towards}. To solve this problem, several methods have been proposed that disrupt the model's functional mapping for samples in $\mathcal{D}_f$ (e.g. by learning random labels), while also training for a small number of epochs over $\mathcal{D}_r$ to repair or protect performance \citep{chundawat2023can, chundawat2023zero, graves2021amnesiac, tarun2023fast, kurmanji2023towards}. This is computationally expensive and requires the storage of the full dataset permanently. In contrast, recent methods such as FisherForgetting from \citet{Golatkar_2020_CVPR} or Selective Synaptic Dampening  (SSD) from \citet{foster2023fast} have explored retraining-free approaches, with the latter achieving state-of-the-art performance. These often involve estimating the importance of parameters for the forget set, allowing them to be disrupted through noise or dampening, but without destroying performance on the retain set. Parameter importance estimation is typically achieved via the diagonal of the Fisher information matrix (FIM), which in practice is calculated as the square of the first-order gradients of the loss. This requires access to a ground truth label, which may not always be available. Therefore, we propose a novel extension of SSD that does not require access to the loss or ground truth labels, which we refer to as Loss-Free Selective Synaptic Dampening (LFSSD). LFSSD replaces the Fisher-based importance estimation with an approximation of the sensitivity of the model to perturbations of each parameter, which may be interpreted as parameter importance. Such importance estimation has been shown to be effective in continual learning \citep{aljundi2018memory}, yet it has not been applied in unlearning. Here, we show this new approach to dampening important parameters is equally effective as Fisher information, without relying on labelled data. To the best of our knowledge, LFSSD is the first retraining-free unlearning method to require only the unlabelled forget samples at forgetting time.\\

\section{Methods \& Results}

\begin{table}[t]
\centering
\caption{Full- (class lamp) and subclass (lamp is part of electrical devices) unlearning with $\alpha=$10, $\lambda=$1 on Cifar100/20. (LF)SSD is deterministic, '$\pm$0.0' omitted. Full benchmarks in the appendix.\\}
\label{fullclass_results}

\fontsize{9pt}{11pt}\selectfont
\setlength{\tabcolsep}{2.2pt}
\begin{tabular}{ll|cccc|cccc}
                            &                   & \multicolumn{4}{c|}{ResNet18}                                                                                                                                                   & \multicolumn{4}{c}{Vision Transformer}                                                                                        \\ \hline
\multicolumn{1}{l|}{data}  & metric            & \multicolumn{1}{c|}{baseline}       & \multicolumn{1}{c|}{retrain}                                   & \multicolumn{1}{c|}{LFSSD}      & SSD                                     & \multicolumn{1}{c|}{baseline}       & \multicolumn{1}{c|}{retrain}        & \multicolumn{1}{c|}{LFSSD}  & SSD                  \\ \hline


 \hline 

\multicolumn{1}{l|}{Cifar100}   & $\mathcal{D}_{r}$ & \multicolumn{1}{c|}{76.39$\pm$0.00} & \multicolumn{1}{c|}{72.89$\pm$0.34}                            & \multicolumn{1}{c|}{76.36}     & 76.08                          & \multicolumn{1}{c|}{88.84$\pm$0.00} & \multicolumn{1}{c|}{90.10$\pm$0.19} & \multicolumn{1}{c|}{89.03} & 89.06       \\
\multicolumn{1}{l|}{}       & $\mathcal{D}_{f}$ & \multicolumn{1}{c|}{70.49$\pm$0.00} & \multicolumn{1}{c|}{0.00$\pm$0.00}                             & \multicolumn{1}{c|}{0.00}         & 0.00                           & \multicolumn{1}{c|}{97.22$\pm$0.00} & \multicolumn{1}{c|}{0.00$\pm$0.00}  & \multicolumn{1}{c|}{0.00}     & 36.89       \\
\multicolumn{1}{l|}{}       & MIA               & \multicolumn{1}{c|}{92.40$\pm$0.00} & \multicolumn{1}{c|}{0.32$\pm$0.00}                             & \multicolumn{1}{c|}{0.00}         & 0.20                           & \multicolumn{1}{c|}{95.60$\pm$0.00} & \multicolumn{1}{c|}{2.27$\pm$0.01}  & \multicolumn{1}{c|}{2.20}   & 0.40        \\ \hline

\multicolumn{1}{l|}{Cifar20}  & $\mathcal{D}_{r}$ & \multicolumn{1}{c|}{82.61$\pm$0.00} & \multicolumn{1}{c|}{81.81$\pm$0.19} & \multicolumn{1}{c|}{81.54}     & 79.58 & \multicolumn{1}{c|}{95.77$\pm$0.00} & \multicolumn{1}{c|}{94.69$\pm$0.13} & \multicolumn{1}{c|}{95.68}     & 95.54 \\
\multicolumn{1}{l|}{}      & $\mathcal{D}_{f}$ & \multicolumn{1}{c|}{72.66$\pm$0.00} & \multicolumn{1}{c|}{22.95$\pm$4.72} & \multicolumn{1}{c|}{7.29}     & 0.00  & \multicolumn{1}{c|}{89.58$\pm$0.00} & \multicolumn{1}{c|}{34.55$\pm$8.62} & \multicolumn{1}{c|}{24.05}     & 14.58 \\
\multicolumn{1}{l|}{}      & MIA               & \multicolumn{1}{c|}{83.40$\pm$0.00} & \multicolumn{1}{c|}{4.29$\pm$0.01}  & \multicolumn{1}{c|}{3.40}     & 2.00  & \multicolumn{1}{c|}{81.00$\pm$0.00} & \multicolumn{1}{c|}{5.60$\pm$0.02}  & \multicolumn{1}{c|}{5.80}     & 3.2   \\ \hline
\end{tabular}
\end{table}

We build upon the approach presented in SSD \citep{foster2023fast}. Given the parameter importance over both $\mathcal{D}$ and $\mathcal{D}_f$, we can select parameters that are disproportionately important for the forget set, and reduce their magnitude to induce forgetting in the model:
\begin{equation}    
    \theta_{i} = 
        \begin{cases}    
            \beta\theta_{i}, & \text{if } []_{\mathcal{D}_{f,i}} > \alpha[]_{\mathcal{D},i}\\
            \theta_{i}, & \text{if } []_{\mathcal{D}_{f,i}} \leq \alpha[]_{\mathcal{D},i}  
    \end{cases}\quad
    \forall i \in [0,|\theta|]\quad\quad
    \text{where } \beta = min(\frac{\lambda []_{\mathcal{D},i}}{[]_{\mathcal{D}_{f,i}}}, 1)
\label{eq:select}
\end{equation}
Here, $\alpha$ and $\lambda$ are hyper-parameters, and $[]_{\mathcal{D}_{f,i}}$ and  $[]_{\mathcal{D}_{i}}$ are the $ith$ element of the diagonal of the FIM, calculated over $\mathcal{D}_f$ and $\mathcal{D}$, respectively. Fisher information is calculated over $\mathcal{D}$ not $\mathcal{D}_{r}$, meaning this is only calculated once, and then the train-set may be discarded after. We replace the Fisher information with the sensitivity estimation from \citet{aljundi2018memory}, which we now introduce. Given some neural network output, $f(x; \theta)$, and some small perturbation $\delta = \{\delta_{i}\}$ in the parameters $\theta=\{\theta_i\}$, we can approximate the change in the neural network's output by:
\begin{equation}
    f(x; \theta + \delta) - f(x; \theta) \approx \sum_{i} \frac{\partial f(x;\theta)}{\partial \theta_{i}} \delta_{i}
\end{equation}
The importance of a parameter can be measured by how much $f(x; \theta)$ changes, given a small perturbation to the parameter. Assuming $\delta_{i}$ is a small, constant change, this is equivalent to the magnitude of the gradient. For multi-dimensional outputs, this  gradient must be calculated over every dimension which is expensive, and so instead the gradient can be calculated over the squared $l_{2}$ norm of the output. Putting this together, we arrive at the final importance estimation equation:
\begin{equation}
    \Omega_{i} = \frac{1}{N} \sum_{k=1}^{N} \lVert \frac{\partial [l_{2}^{2} (f(x_k; \theta))]}{\partial \theta_{i}} \rVert
\end{equation}
Which we can calculate over $\mathcal{D}$ and $\mathcal{D}_f$ and substitute into equation \ref{eq:select} in place of $[]$ and $[]_{f}$. Results can be seen in Table \ref{fullclass_results}, with further results available in the appendix. The objective is to maximise the $\mathcal{D}_{r}$ accuracy while having a low membership inference attack score (MIA) similar to the retrain column \citep{shokri2017membership}. 
Similarity to the retrained model is necessary, as "over-forgetting" can lead to the \textit{Streisand effect} (e.g., labelling a cat as a boat is an obvious modification) \citep{chundawat2023can}.
The results show that LFSSD produces scores similar to that of a model trained on only $\mathcal{D}_r$, and is competitive with state-of-the-art methods such as SSD, while being the first method to require no labelled data and no retraining. A shared limitation with SSD is the reliance on choosing $\alpha$ and $\lambda$ values that balance forgetting and model performance.

\section{Conclusion}
Developing lightweight yet effective machine unlearning approaches is a significant open problem. In this work, we presented a novel extension of existing literature that is the first retraining free machine unlearning method to require only the unlabelled input samples for the forget set. By not requiring the long-term storage of the retain dataset, a labelled forget set, or the loss over the forget set, LFSSD is lightweight, robust and more practically useful than its predecessors.

\section*{URM Statement}
The authors acknowledge that at least one key author of this work meets the URM criteria of ICLR 2024 Tiny Papers Track

\bibliography{iclr2024_conference}
\bibliographystyle{iclr2024_conference}

\appendix
\section{Appendix}


\begin{figure}[h]
\begin{center}
\includegraphics[width=1.0\textwidth]{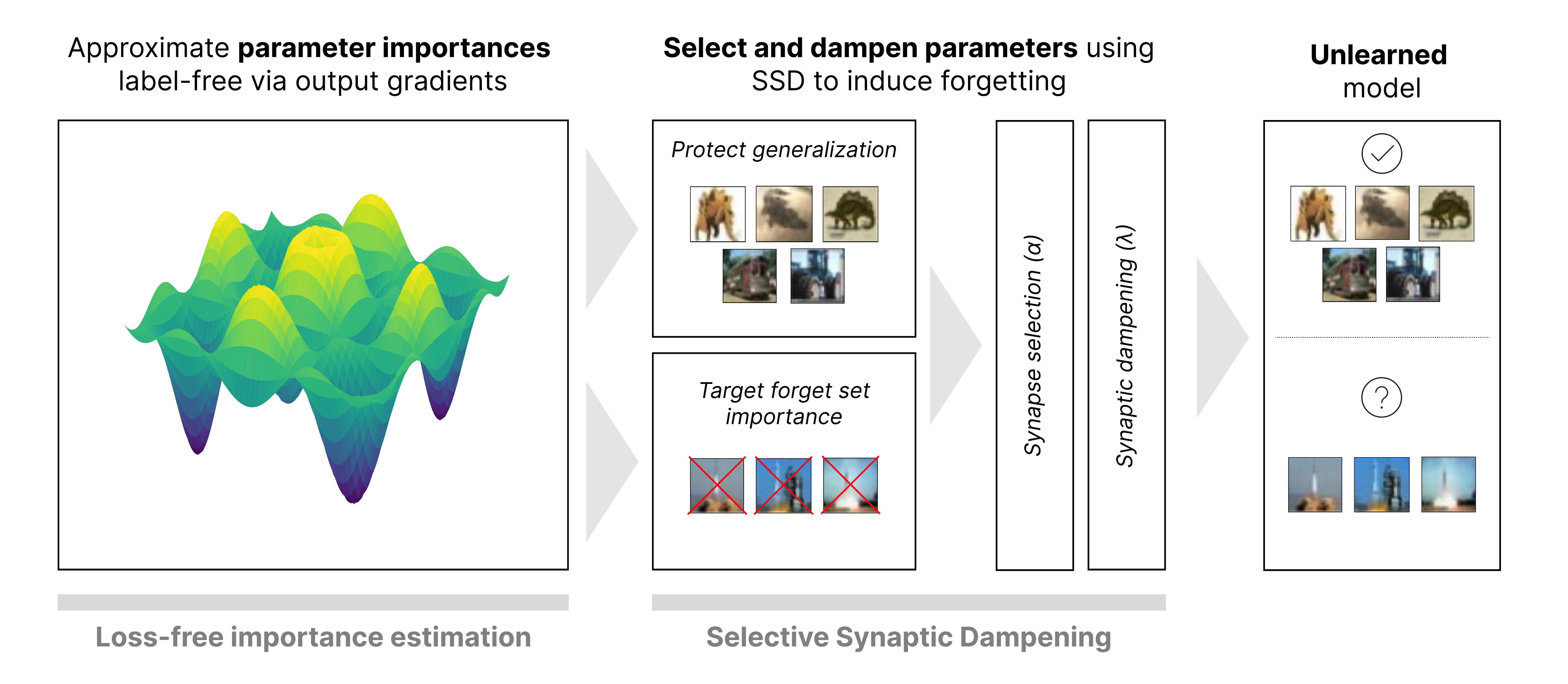}
\end{center}
\caption{Loss-free parameter importance estimation as a replacement for the label-dependent parameter importance estimation of SSD resulting in the LFSSD method.}
\end{figure}

\textbf{Benchmarks.} We evaluate LFSSD in the same manner as \citet{chundawat2023can, foster2023fast}, comparing against a baseline model without any unlearning, a model retrained from scratch on only $\mathcal{D}_r$, a model finetuned for 2 epochs on  $\mathcal{D}_r$, and several existing state-of-the-art methods \citep{chundawat2023can, graves2021amnesiac, foster2023fast} (teacher, amnesiac, SSD). Three key tasks are evaluated using ResNet18 (RN) and Vision Transformer (ViT): full class, subclass, and random forgetting. Fullclass entails forgetting an entire class from the dataset (e.g. rocket from Cifar100), subclass involves forgetting a related subclass from the dataset (e.g. all rockets from class Vehicle2 in Cifar20), and random forgetting selects a random subset of samples across all classes. Datasets used include the Cifar suite \citep{CIFAR_krizhevsky2010convolutional}, and the PINS face recognition dataset \citep{pinsds}. Runs are repeated 10 times for each method, with the mean and standard deviation shown in the tables. $D_r$ and $D_f$ in the tables denote the accuracy on the respective subset of data in percent. Since SSD (and therefore LFSSD) is deterministic, there is no standard deviation across repeated runs. We use the same membership inference attack as in \citet{chundawat2023can, foster2023fast}, where logistic regression is used to classify samples as seen or unseen for a given model, based the entropy of the model outputs for that sample.

\textbf{Resources.} A single backward pass over the full dataset $D$ is $O(b_{full} \cdot p)$ where $b_{full}$ is the number of batches in the dataset and $p$ is the number of parameters in the model. Similarly, the complexity for the forget and retain set is $O(b_{forget} \cdot p)$ and $O(b_{retain} \cdot p)$, respectively. Note that $b_{retain} \approx b_{full} >> b_{forget}$. SSD requires only one computation of the gradients for $D$ (can be saved and reused), then for every forget request it requires one computation for $D_f$ to then apply the dampening in $O(b_{forget} \cdot p)$. 

LFSSD has the same complexity as SSD with the difference being that LFSSD calculates the norm of the output instead of the loss. In practice, this is similar in compute time.

Other methods are retraining-based and depend on the number of epochs $N$, generally on the order of $O(N \cdot b_{retain} \cdot p)$, which is much larger. 

\textbf{Sensitivity analysis}. The values for alpha and lambda for SSD are taken from \citet{foster2023fast} where the authors used Optuna to search for ideal parameters. For LFSSD we performed a simplified hyperparameter search keeping lambda fixed at 1 and alpha values [1, 50]. We pick one value per unlearning task (e.g. full class, sub-class, random) to show robustness across classes and do not use the best alpha per individual class/subclass (e.g., taking alpha 25 for rocket, alpha 5 for another class, etc.). A sensitivity analysis in Table \ref{tab:sensitivity} on CIFAR100 rocket class unlearning shows that there is a significant drop off in performance when picking an alpha that is too aggressive or conservative but the method has a wide plateau in which it performs well.

\textbf{Unlearning scenarios without labels.} For cases where labels are hard to get we expect two main scenarios: First, having a catalogue of undesired/copyrighted data (e.g., images) that we want to unlearn from the model just to be on the safe side legally even if it might not be in the model. In this case, we do not have the labels. Second, models that were trained with semi-supervised learning could benefit from our method when we do not want to rely on the accuracy of pseudo labels.

\begin{table}[]
\fontsize{9pt}{12pt}\selectfont
\setlength{\tabcolsep}{1.3pt}
\caption{Sensitivity analysis for LFSSD parameter $\alpha$ with $\lambda = 1$ on CIFAR100 rocket class unlearning. LFSSD is deterministic, thus no $\pm$ reported.\\}
\centering
\begin{tabular}{c|c|c|c}
\hline
$\alpha$ & $D_r$ Accuracy & $D_f$  Accuracy & MIA   \\ \hline
1                     & 16.87\%        & 0.00\%            & 26.60 \\ \hline
5                     & 74.72\%        & 0.00\%            & 0.00  \\ \hline
10                    & 75.67\%        & 0.00\%            & 0.00  \\ \hline
25                    & 76.10\%          & 0.00\%            & 0.00  \\ \hline
50                    & 76.32\%          & 71.88\%           & 29.40 \\ \hline
\end{tabular}
\label{tab:sensitivity}
\end{table}

\begin{table}[]
\fontsize{9pt}{12pt}\selectfont
\setlength{\tabcolsep}{1.3pt}
\caption{Face unlearning (class unlearning): One face unlearned per experiment [ID 1,10,20,30,40] and results aggregated for 5 experiments as in \citet{foster2023fast}. SSD: $\alpha=50$, $\lambda=0.1$;  LFSSD: $\alpha=10$, $\lambda=1$\\}
\centering
\begin{tabular}{ll|c|c|c|c|c|c|c}
\hline
   & metric            & baseline       & retrain                                   & finetune       & teacher                                  & amnesiac                                  & SSD                                     & LFSSD \\ \hline
RN & $\mathcal{D}_{r}$ & 98.52$\pm$0.02 & 100.00$\pm$0.00 & 99.72$\pm$0.45 & 96.72$\pm$0.44                           & 99.99$\pm$0.02  & 98.42$\pm$0.13                          &    98.15$\pm$0.20   \\
   & $\mathcal{D}_{f}$ & 97.84$\pm$1.99 &  0.00$\pm$0.00  & 4.32$\pm$4.61  & 0.13$\pm$0.4                             &  0.00$\pm$0.00  & 0.00$\pm$0.00 &  0.00$\pm$0.00     \\
   & MIA               & 34.38$\pm$0.23 & 0.00$\pm$0.00   & 0.80$\pm$0.01  &  0.02$\pm$0.00 & 8.92$\pm$0.03                             & 1.11$\pm$0.01                           &    0.38$\pm$0.58   \\ \hline
\end{tabular}
\end{table}

\begin{table}[]

\centering
\caption{Class unlearning on Cifar100. SSD: $\alpha_{RN}=10$, $\alpha_{ViT}=10$, $\lambda=1$;  LFSSD: $\alpha=10$, $\lambda=1$. MR: mushroom, Rkt: Rocket.\\}

\fontsize{9pt}{12pt}\selectfont
\setlength{\tabcolsep}{1.3pt}
\begin{tabular}{ll|l|c|c|c|c|c|c|c}
\hline
    & class & metric            & baseline       & retrain        & finetune       & teacher       & amnesiac       & SSD             & LFSSD \\ \hline
RN  & baby  & $\mathcal{D}_{r}$ & 76.38$\pm$0.00 & 73.10$\pm$0.55 & 64.03$\pm$0.8  & 74.71$\pm$0.19  & 73.69$\pm$0.27 & 44.05*$\pm$0.00 &   75.80$\pm$0.00    \\
    &       & $\mathcal{D}_{f}$ & 72.48$\pm$0.00 & 0.00$\pm$0.00  & 0.00$\pm$0.00  & 0.08$\pm$0.25   & 0.00$\pm$0.00  & 0.00$\pm$0.00   &   0.00$\pm$0.00    \\
    &       & MIA               & 92.60$\pm$0.00 & 2.44$\pm$0.01  & 24.52$\pm$0.07 & 0.00$\pm$0.00   & 50.86$\pm$0.05 & 5.40$\pm$0.00   &    0.00$\pm$0.00   \\ \cline{2-10} 
    & lamp  & $\mathcal{D}_{r}$ & 76.39$\pm$0.00 & 72.89$\pm$0.34 & 64.01$\pm$0.63 & 74.76$\pm$0.19  & 73.52$\pm$0.48 & 76.08$\pm$0.00  &   76.36$\pm$0.00    \\
    &       & $\mathcal{D}_{f}$ & 70.49$\pm$0.00 & 0.00$\pm$0.00  & 0.00$\pm$0.00  & 0.00$\pm$0.00   & 0.00$\pm$0.00  & 0.00$\pm$0.00   &  0.00$\pm$0.00     \\
    &       & MIA               & 92.40$\pm$0.00 & 0.32$\pm$0.00  & 12.92$\pm$0.04 & 0.00$\pm$0.00   & 46.24$\pm$0.04 & 0.20$\pm$0.00   &    0.00$\pm$0.00   \\ \cline{2-10} 
    & MR    & $\mathcal{D}_{r}$ & 76.28$\pm$0.00 & 72.90$\pm$0.45 & 63.97$\pm$0.67 & 74.53$\pm$0.26  & 73.56$\pm$0.48 & 75.59$\pm$0.00  &    76.28$\pm$0.00   \\
    &       & $\mathcal{D}_{f}$ & 80.12$\pm$0.00 & 0.00$\pm$0.00  & 0.00$\pm$0.00  & 0.00$\pm$0.00   & 0.00$\pm$0.00  & 0.00$\pm$0.00   &     0.00$\pm$0.00  \\
    &       & MIA               & 95.20$\pm$0.00 & 0.22$\pm$0.00  & 12.98$\pm$0.03 & 0.00$\pm$0.00   & 46.48$\pm$0.04 & 0.20$\pm$0.00   &    0.00$\pm$0.00   \\ \cline{2-10} 
    & Rkt   & $\mathcal{D}_{r}$ & 76.27$\pm$0.00 & 72.83$\pm$0.42 & 64.05$\pm$0.88 & 74.53$\pm$0.26  & 73.34$\pm$0.45 & 74.54$\pm$0.00  &   75.67$\pm$0.00    \\
    &       & $\mathcal{D}_{f}$ & 80.90$\pm$0.00 & 0.00$\pm$0.00  & 0.00$\pm$0.00  & 0.00$\pm$0.00   & 0.00$\pm$0.00  & 0.00$\pm$0.00   &   0.00$\pm$0.00    \\
    &       & MIA               & 93.40$\pm$0.00 & 1.04$\pm$0.00  & 13.70$\pm$0.04 & 0.00$\pm$0.00   & 29.56$\pm$0.02 & 2.20$\pm$0.00   &   0.00$\pm$0.00    \\ \cline{2-10} 
    & sea   & $\mathcal{D}_{r}$ & 76.23$\pm$0.00 & 72.83$\pm$0.54 & 63.80$\pm$1.36 & 74.56$\pm$0.18  & 73.14$\pm$0.42 & 73.56$\pm$0.00  &   75.43$\pm$0.00    \\
    &       & $\mathcal{D}_{f}$ & 85.85$\pm$0.00 & 0.00$\pm$0.00  & 0.00$\pm$0.00  & 0.08$\pm$0.25   & 0.00$\pm$0.00  & 0.00$\pm$0.00   &  0.00$\pm$0.00     \\
    &       & MIA               & 93.40$\pm$0.00 & 5.84$\pm$0.02  & 26.54$\pm$0.08 & 0.02$\pm$0.00   & 29.50$\pm$0.03 & 0.60$\pm$0.00   &  0.00$\pm$0.00     \\ \hline
ViT & baby  & $\mathcal{D}_{r}$ & 88.93$\pm$0.00 & 90.27$\pm$0.15 & 80.74$\pm$1.38 & 87.48$\pm$0.41  & 88.43$\pm$0.71 & 88.59$\pm$0.00  &   88.12$\pm$0.00    \\
    &       & $\mathcal{D}_{f}$ & 90.19$\pm$0.00 & 0.00$\pm$0.00  & 0.00$\pm$0.00  & 23.80$\pm$22.51 & 0.00$\pm$0.00  & 0.00$\pm$0.00   &    0.00$\pm$0.00   \\
    &       & MIA               & 75.60$\pm$0.00 & 21.53$\pm$0.03 & 26.77$\pm$0.13 & 0.00$\pm$0.00   & 1.83$\pm$0.00  & 0.60$\pm$0.00   &   4.00$\pm$0.00    \\ \cline{2-10} 
    & lamp  & $\mathcal{D}_{r}$ & 88.84$\pm$0.00 & 90.10$\pm$0.19 & 80.25$\pm$1.48 & 87.50$\pm$0.43  & 88.43$\pm$0.6  & 89.06$\pm$0.00  &  89.03$\pm$0.00     \\
    &       & $\mathcal{D}_{f}$ & 97.22$\pm$0.00 & 0.00$\pm$0.00  & 0.36$\pm$0.89  & 25.25$\pm$12.5  & 0.00$\pm$0.00  & 36.89$\pm$0.00  &  0.00$\pm$0.00     \\
    &       & MIA               & 95.60$\pm$0.00 & 2.27$\pm$0.01  & 11.77$\pm$0.04 & 0.13$\pm$0.00   & 2.70$\pm$0.00  & 0.40$\pm$0.00   &  2.20$\pm$0.00     \\ \cline{2-10} 
    & MR    & $\mathcal{D}_{r}$ & 88.87$\pm$0.00 & 90.02$\pm$0.22 & 81.14$\pm$0.79 & 87.42$\pm$0.41  & 88.34$\pm$0.72 & 88.82$\pm$0.00  &  88.93$\pm$0.00     \\
    &       & $\mathcal{D}_{f}$ & 94.88$\pm$0.00 & 0.00$\pm$0.00  & 2.33$\pm$2.37  & 12.82$\pm$5.92  & 0.00$\pm$0.00  & 0.00$\pm$0.00   &   0.00$\pm$0.00    \\
    &       & MIA               & 92.80$\pm$0.00 & 0.70$\pm$0.00  & 7.10$\pm$0.02  & 0.03$\pm$0.00   & 0.47$\pm$0.00  & 3.80$\pm$0.00   &   2.60$\pm$0.00    \\ \cline{2-10} 
    & Rkt   & $\mathcal{D}_{r}$ & 88.88$\pm$0.00 & 90.07$\pm$0.09 & 80.82$\pm$1.37 & 87.46$\pm$0.53  & 87.92$\pm$0.89 & 88.90$\pm$0.00  &   88.86$\pm$0.00    \\
    &       & $\mathcal{D}_{f}$ & 94.70$\pm$0.00 & 0.00$\pm$0.00  & 0.46$\pm$0.72  & 4.20$\pm$5.24   & 0.00$\pm$0.00  & 0.00$\pm$0.00   &  0.00$\pm$0.00     \\
    &       & MIA               & 94.40$\pm$0.00 & 3.23$\pm$0.00  & 19.00$\pm$0.09 & 0.03$\pm$0.00   & 1.00$\pm$0.01  & 1.80$\pm$0.00   &  10.60$\pm$0.00     \\ \cline{2-10} 
    & sea   & $\mathcal{D}_{r}$ & 88.91$\pm$0.00 & 90.27$\pm$0.21 & 80.82$\pm$1.36 & 87.72$\pm$0.22  & 88.25$\pm$0.29 & 87.95$\pm$0.00  &   88.67$\pm$0.00    \\
    &       & $\mathcal{D}_{f}$ & 90.54$\pm$0.00 & 0.00$\pm$0.00  & 0.00$\pm$0.00  & 51.13$\pm$17.37 & 0.00$\pm$0.00  & 0.00$\pm$0.00   &   0.00$\pm$0.00    \\
    &       & MIA               & 80.40$\pm$0.00 & 8.43$\pm$0.02  & 21.97$\pm$0.07 & 0.00$\pm$0.00   & 0.77$\pm$0.00  & 3.20$\pm$0.00   &  4.20$\pm$0.00     \\ \hline
\end{tabular}
\end{table}

\begin{table}[]
\fontsize{9pt}{12pt}\selectfont
\setlength{\tabcolsep}{1.3pt}
\caption{Class unlearning on Cifar20. SSD: $\alpha_{RN}=10$, $\alpha_{ViT}=10$, $\lambda=1$;  LFSSD: $\alpha=5$, $\lambda=1$. ED: electrical devices, NS: natural scenes, veg: vegetables, Veh2: vehicle2.\\}
\centering
\begin{tabular}{ll|l|c|c|c|c|c|c|c}
\hline
    & class  & metric            & baseline       & retrain        & finetune        & teacher      & amnesiac       & SSD            & LFSSD \\ \hline
RN  & ED     & $\mathcal{D}_{r}$ & 82.56$\pm$0.00 & 82.13$\pm$0.23 & 73.19$\pm$1.08  & 82.04$\pm$0.29 & 81.34$\pm$0.18 & 83.15$\pm$0.00 &  83.19$\pm$0.00   \\
    &        & $\mathcal{D}_{f}$ & 82.26$\pm$0.00 & 0.00$\pm$0.00  & 0.00$\pm$0.00   & 10.88$\pm$1.3  & 0.00$\pm$0.00  & 1.76$\pm$0.00  &   0.00$\pm$0.00  \\
    &        & MIA               & 89.56$\pm$0.00 & 8.91$\pm$0.01  & 30.37$\pm$0.05  & 0.00$\pm$0.00  & 7.02$\pm$0.01  & 4.16$\pm$0.00  &  3.80$\pm$0.00   \\ \cline{2-10} 
    & NS     & $\mathcal{D}_{r}$ & 82.10$\pm$0.00 & 81.33$\pm$0.22 & 72.03$\pm$1.63  & 81.36$\pm$0.27 & 80.70$\pm$0.4  & 82.33$\pm$0.00 &  82.44$\pm$0.00   \\
    &        & $\mathcal{D}_{f}$ & 91.08$\pm$0.00 & 0.00$\pm$0.00  & 0.00$\pm$0.00   & 10.91$\pm$2.16 & 0.00$\pm$0.00  & 0.00$\pm$0.00  &  0.00$\pm$0.00   \\
    &        & MIA               & 88.68$\pm$0.00 & 3.77$\pm$0.01  & 17.60$\pm$0.03  & 0.00$\pm$0.00  & 3.71$\pm$0.01  & 3.28$\pm$0.00  &  2.68$\pm$0.00   \\ \cline{2-10} 
    & people & $\mathcal{D}_{r}$ & 82.11$\pm$0.00 & 81.20$\pm$0.19 & 72.46$\pm$1.01  & 81.31$\pm$0.16 & 80.64$\pm$0.34 & 82.31$\pm$0.00 &  82.08$\pm$0.00   \\
    &        & $\mathcal{D}_{f}$ & 90.70$\pm$0.00 & 0.00$\pm$0.00  & 0.00$\pm$0.00   & 1.07$\pm$0.48  & 0.00$\pm$0.00  & 0.00$\pm$0.00  &  0.00$\pm$0.00   \\
    &        & MIA               & 91.72$\pm$0.00 & 1.36$\pm$0.00  & 16.20$\pm$0.04  & 0.00$\pm$0.00  & 6.28$\pm$0.01  & 1.12$\pm$0.00  &  1.36$\pm$0.00   \\ \cline{2-10} 
    & veg    & $\mathcal{D}_{r}$ & 82.31$\pm$0.00 & 81.39$\pm$0.21 & 71.42$\pm$1.32  & 81.46$\pm$0.3  & 81.01$\pm$0.33 & 82.38$\pm$0.00 &  82.47$\pm$0.00   \\
    &        & $\mathcal{D}_{f}$ & 86.90$\pm$0.00 & 0.00$\pm$0.00  & 0.00$\pm$0.00   & 2.67$\pm$1.35  & 0.00$\pm$0.00  & 0.00$\pm$0.00  &  0.00$\pm$0.00   \\
    &        & MIA               & 89.52$\pm$0.00 & 9.74$\pm$0.01  & 29.39$\pm$0.08  & 0.00$\pm$0.00  & 5.00$\pm$0.01  & 16.96$\pm$0.00 &  5.44$\pm$0.00   \\ \cline{2-10} 
    & Veh2   & $\mathcal{D}_{r}$ & 82.69$\pm$0.00 & 82.11$\pm$0.19 & 73.50$\pm$0.86  & 81.96$\pm$0.21 & 81.13$\pm$0.3  & 82.97$\pm$0.00 &  82.72$\pm$0.00   \\
    &        & $\mathcal{D}_{f}$ & 80.41$\pm$0.00 & 0.00$\pm$0.00  & 0.00$\pm$0.00   & 3.62$\pm$1.07  & 0.00$\pm$0.00  & 0.00$\pm$0.00  &  0.00$\pm$0.00   \\
    &        & MIA               & 82.56$\pm$0.00 & 13.54$\pm$0.01 & 30.63$\pm$0.04  & 0.00$\pm$0.00  & 7.54$\pm$0.01  & 6.68$\pm$0.00  &  3.92$\pm$0.00   \\ \hline
ViT & ED     & $\mathcal{D}_{r}$ & 95.73$\pm$0.00 & 94.71$\pm$0.14 & 87.35$\pm$1.44  & 93.42$\pm$0.62 & 93.45$\pm$0.44 & 95.82$\pm$0.00 &   95.84$\pm$0.00  \\
    &        & $\mathcal{D}_{f}$ & 95.03$\pm$0.00 & 0.00$\pm$0.00  & 0.28$\pm$0.22   & 4.14$\pm$3.75  & 0.03$\pm$0.08  & 53.53$\pm$0.00 &   0.68$\pm$0.00  \\
    &        & MIA               & 91.60$\pm$0.00 & 9.82$\pm$0.01  & 23.60$\pm$0.04  & 0.02$\pm$0.00  & 1.70$\pm$0.00  & 1.32$\pm$0.00  &   4.00$\pm$0.00  \\ \cline{2-10} 
    & NS     & $\mathcal{D}_{r}$ & 95.71$\pm$0.00 & 94.79$\pm$0.11 & 87.45$\pm$1.2   & 93.50$\pm$0.4  & 93.68$\pm$0.56 & 93.63$\pm$0.00 &  95.13$\pm$0.00   \\
    &        & $\mathcal{D}_{f}$ & 95.37$\pm$0.00 & 0.00$\pm$0.00  & 0.05$\pm$0.15   & 2.63$\pm$2.08  & 0.00$\pm$0.00  & 0.00$\pm$0.00  &  0.00$\pm$0.00   \\
    &        & MIA               & 85.04$\pm$0.00 & 4.70$\pm$0.01  & 16.97$\pm$0.04  & 0.06$\pm$0.00  & 1.04$\pm$0.00  & 1.88$\pm$0.00  &  4.72$\pm$0.00   \\ \cline{2-10} 
    & people & $\mathcal{D}_{r}$ & 95.54$\pm$0.00 & 94.54$\pm$0.14 & 80.18$\pm$19.88 & 93.19$\pm$0.54 & 93.28$\pm$0.41 & 95.33$\pm$0.00 &  95.50$\pm$0.00   \\
    &        & $\mathcal{D}_{f}$ & 98.54$\pm$0.00 & 0.00$\pm$0.00  & 0.09$\pm$0.14   & 2.91$\pm$3.36  & 0.00$\pm$0.00  & 0.00$\pm$0.00  &  2.52$\pm$0.00   \\
    &        & MIA               & 89.48$\pm$0.00 & 1.56$\pm$0.00  & 8.08$\pm$0.03   & 0.01$\pm$0.00  & 0.60$\pm$0.00  & 1.20$\pm$0.00  &  1.00$\pm$0.00   \\ \cline{2-10} 
    & veg    & $\mathcal{D}_{r}$ & 95.59$\pm$0.00 & 94.54$\pm$0.21 & 87.09$\pm$1.24  & 92.92$\pm$0.51 & 93.29$\pm$0.41 & 95.71$\pm$0.00 &  95.64$\pm$0.00   \\
    &        & $\mathcal{D}_{f}$ & 97.57$\pm$0.00 & 0.00$\pm$0.00  & 0.30$\pm$0.29   & 8.28$\pm$6.79  & 0.02$\pm$0.07  & 0.00$\pm$0.00  &  0.59$\pm$0.00   \\
    &        & MIA               & 91.32$\pm$0.00 & 4.41$\pm$0.01  & 14.72$\pm$0.05  & 0.02$\pm$0.00  & 1.02$\pm$0.00  & 1.88$\pm$0.00  &  3.00$\pm$0.00   \\ \cline{2-10} 
    & Veh2   & $\mathcal{D}_{r}$ & 95.73$\pm$0.00 & 94.85$\pm$0.13 & 87.75$\pm$1.64  & 93.59$\pm$0.3  & 93.88$\pm$0.15 & 93.12$\pm$0.00 &  95.94$\pm$0.00   \\
    &        & $\mathcal{D}_{f}$ & 95.22$\pm$0.00 & 0.00$\pm$0.00  & 0.04$\pm$0.12   & 4.88$\pm$4.12  & 0.00$\pm$0.00  & 0.00$\pm$0.00  &  0.00$\pm$0.00   \\
    &        & MIA               & 84.04$\pm$0.00 & 22.96$\pm$0.03 & 38.15$\pm$0.08  & 0.02$\pm$0.00  & 1.20$\pm$0.00  & 7.04$\pm$0.00  &  9.12$\pm$0.00   \\ \hline
\end{tabular}
\end{table}

\begin{table}[]
\fontsize{9pt}{12pt}\selectfont
\setlength{\tabcolsep}{1.3pt}
\caption{Subclass unlearning on Cifar20. SSD: $\alpha_{RN}=10$, $\alpha_{ViT}=25$, $\lambda=1$;  LFSSD: $\alpha=10$, $\lambda=1$. MR: mushroom, Rkt: Rocket.\\}
\centering
\begin{tabular}{ll|l|c|c|c|c|c|c|c}
\hline
    & class & metric            & baseline       & retrain        & finetune        & teacher       & amnesiac       & SSD            & LFSSD \\ \hline
RN  & baby  & $\mathcal{D}_{r}$ & 82.45$\pm$0.00 & 81.45$\pm$0.3  & 72.94$\pm$0.82  & 81.25$\pm$0.19  & 81.02$\pm$0.2  & 79.25$\pm$0.00 &  78.93$\pm$0.00     \\
    &       & $\mathcal{D}_{f}$ & 86.98$\pm$0.00 & 80.03$\pm$3.4  & 66.52$\pm$6.19  & 75.17$\pm$2.66  & 37.70$\pm$9.59 & 8.85$\pm$0.00  &  0.78$\pm$0.00     \\
    &       & MIA               & 90.40$\pm$0.00 & 44.82$\pm$0.02 & 56.20$\pm$0.06  & 0.04$\pm$0.00   & 13.65$\pm$0.02 & 1.40$\pm$0.00  &  1.00$\pm$0.00     \\ \cline{2-10} 
    & lamp  & $\mathcal{D}_{r}$ & 82.61$\pm$0.00 & 81.81$\pm$0.19 & 72.46$\pm$0.95  & 81.51$\pm$0.19  & 81.36$\pm$0.36 & 79.58$\pm$0.00 &  81.54$\pm$0.00     \\
    &       & $\mathcal{D}_{f}$ & 72.66$\pm$0.00 & 22.95$\pm$4.72 & 17.25$\pm$6.58  & 16.82$\pm$3.07  & 7.44$\pm$1.73  & 0.00$\pm$0.00  &  7.29$\pm$0.00     \\
    &       & MIA               & 83.40$\pm$0.00 & 4.29$\pm$0.01  & 21.55$\pm$0.04  & 0.04$\pm$0.00   & 15.73$\pm$0.03 & 2.00$\pm$0.00  &  3.40$\pm$0.00     \\ \cline{2-10} 
    & MR    & $\mathcal{D}_{r}$ & 82.60$\pm$0.00 & 81.66$\pm$0.21 & 72.51$\pm$1.15  & 81.50$\pm$0.2   & 81.44$\pm$0.18 & 80.05$\pm$0.00 &  80.38$\pm$0.00     \\
    &       & $\mathcal{D}_{f}$ & 72.22$\pm$0.00 & 10.48$\pm$2.02 & 7.46$\pm$4.89   & 8.77$\pm$2.79   & 1.09$\pm$1.13  & 0.78$\pm$0.00  &  0.78$\pm$0.00     \\
    &       & MIA               & 86.80$\pm$0.00 & 2.27$\pm$0.00  & 18.09$\pm$0.02  & 0.00$\pm$0.00   & 12.80$\pm$0.03 & 4.00$\pm$0.00  &   3.80$\pm$0.00    \\ \cline{2-10} 
    & Rkt   & $\mathcal{D}_{r}$ & 82.54$\pm$0.00 & 81.54$\pm$0.24 & 72.41$\pm$0.95  & 81.48$\pm$0.27  & 81.46$\pm$0.26 & 82.43$\pm$0.00 &  80.49$\pm$0.00     \\
    &       & $\mathcal{D}_{f}$ & 79.34$\pm$0.00 & 10.74$\pm$3.4  & 9.75$\pm$6.68   & 6.41$\pm$3.57   & 0.76$\pm$0.73  & 2.17$\pm$0.00  &  0.00$\pm$0.00     \\
    &       & MIA               & 89.40$\pm$0.00 & 3.85$\pm$0.01  & 18.67$\pm$0.05  & 0.00$\pm$0.00   & 6.60$\pm$0.01  & 10.80$\pm$0.00 &  6.60$\pm$0.00     \\ \cline{2-10} 
    & sea   & $\mathcal{D}_{r}$ & 82.37$\pm$0.00 & 81.30$\pm$0.27 & 72.50$\pm$1.55  & 81.22$\pm$0.24  & 81.05$\pm$0.31 & 81.72$\pm$0.00 & 79.84$\pm$0.00      \\
    &       & $\mathcal{D}_{f}$ & 96.27$\pm$0.00 & 91.47$\pm$1.92 & 82.69$\pm$7.17  & 75.13$\pm$4.12  & 46.78$\pm$8.55 & 75.35$\pm$0.00 &   18.06$\pm$0.00    \\
    &       & MIA               & 90.80$\pm$0.00 & 52.09$\pm$0.03 & 62.82$\pm$0.11  & 0.00$\pm$0.00   & 4.45$\pm$0.01  & 21.80$\pm$0.00 &  2.20$\pm$0.00     \\ \hline
ViT & baby  & $\mathcal{D}_{r}$ & 95.69$\pm$0.00 & 94.50$\pm$0.19 & 87.60$\pm$0.75  & 92.98$\pm$0.53  & 93.35$\pm$0.26 & 95.54$\pm$0.00 &   95.38$\pm$0.00    \\
    &       & $\mathcal{D}_{f}$ & 96.44$\pm$0.00 & 93.23$\pm$1.09 & 85.45$\pm$4.47  & 46.66$\pm$17.88 & 38.76$\pm$7.36 & 94.10$\pm$0.00 &  86.81$\pm$0.00     \\
    &       & MIA               & 91.60$\pm$0.00 & 77.37$\pm$0.03 & 66.57$\pm$0.07  & 0.03$\pm$0.00   & 0.93$\pm$0.01  & 77.20$\pm$0.00 &   1.60$\pm$0.00    \\ \cline{2-10} 
    & lamp  & $\mathcal{D}_{r}$ & 95.77$\pm$0.00 & 94.69$\pm$0.13 & 87.72$\pm$0.52  & 93.57$\pm$0.67  & 93.66$\pm$0.5  & 95.54$\pm$0.00 &  95.68$\pm$0.00     \\
    &       & $\mathcal{D}_{f}$ & 89.58$\pm$0.00 & 34.55$\pm$8.62 & 16.90$\pm$10.4  & 8.23$\pm$7.05   & 0.59$\pm$1.45  & 14.58$\pm$0.00 &  24.05$\pm$0.00     \\
    &       & MIA               & 81.00$\pm$0.00 & 5.60$\pm$0.02  & 14.73$\pm$0.04  & 0.10$\pm$0.00   & 2.00$\pm$0.01  & 3.2$\pm$0.00   &  5.80$\pm$0.00     \\ \cline{2-10} 
    & MR    & $\mathcal{D}_{r}$ & 95.69$\pm$0.00 & 94.60$\pm$0.13 & 87.37$\pm$0.91  & 93.55$\pm$0.42  & 93.42$\pm$0.46 & 95.51$\pm$0.00 &  95.66$\pm$0.00     \\
    &       & $\mathcal{D}_{f}$ & 97.05$\pm$0.00 & 26.57$\pm$6.41 & 15.71$\pm$12.13 & 13.01$\pm$9.11  & 0.20$\pm$0.36  & 6.68$\pm$0.00  &  31.42$\pm$0.00     \\
    &       & MIA               & 77.80$\pm$0.00 & 2.34$\pm$0.01  & 9.25$\pm$0.04   & 0.00$\pm$0.00   & 1.50$\pm$0.01  & 0.40$\pm$0.00  &  0.80$\pm$0.00     \\ \cline{2-10} 
    & Rkt   & $\mathcal{D}_{r}$ & 95.73$\pm$0.00 & 94.61$\pm$0.13 & 85.70$\pm$3.05  & 93.60$\pm$0.29  & 93.47$\pm$0.22 & 95.13$\pm$0.00 &  95.66$\pm$0.00     \\
    &       & $\mathcal{D}_{f}$ & 94.53$\pm$0.00 & 22.26$\pm$8.34 & 6.25$\pm$6.03   & 3.35$\pm$2.89   & 0.85$\pm$1.71  & 5.12$\pm$0.00  &  2.17$\pm$0.00     \\
    &       & MIA               & 80.40$\pm$0.00 & 3.44$\pm$0.01  & 16.04$\pm$0.03  & 0.02$\pm$0.00   & 0.78$\pm$0.00  & 5.40$\pm$0.00  &  3.00$\pm$0.00     \\ \cline{2-10} 
    & sea   & $\mathcal{D}_{r}$ & 95.67$\pm$0.00 & 94.55$\pm$0.22 & 87.65$\pm$1.56  & 93.57$\pm$0.26  & 93.26$\pm$0.24 & 95.57$\pm$0.00 &  94.45$\pm$0.00     \\
    &       & $\mathcal{D}_{f}$ & 99.22$\pm$0.00 & 95.12$\pm$0.81 & 89.17$\pm$4.17  & 25.97$\pm$14.01 & 21.42$\pm$8.5  & 97.05$\pm$0.00 &  20.83$\pm$0.00     \\
    &       & MIA               & 88.40$\pm$0.00 & 65.96$\pm$0.04 & 65.04$\pm$0.13  & 0.17$\pm$0.00   & 0.40$\pm$0.00  & 82.20$\pm$0.00 &  6.60$\pm$0.00     \\ \hline
\end{tabular}
\end{table}

\begin{table}[]
\fontsize{9pt}{12pt}\selectfont
\setlength{\tabcolsep}{1.3pt}
\caption{Random sample unlearning on Cifar10 (100 samples). SSD: $\alpha=10$, $\lambda=1$;  LFSSD: $\alpha=3.5$, $\lambda=1$\\}
\centering
\begin{tabular}{ll|l|c|c|c|c|c|c}
\hline
    & metric            & baseline                                 & retrain                                  & finetune       & teacher                                  & amnesiac       & \multicolumn{1}{l|}{SSD}                 & \multicolumn{1}{l}{LFSSD} \\ \hline
RN  & $\mathcal{D}_{r}$ & 90.71$\pm$0.00                           & 91.45$\pm$0.11 & 88.02$\pm$0.45 & 90.21$\pm$0.10 & 90.16$\pm$0.23 & 88.68$\pm$3.36                           &          89.82$\pm$2.30              \\
    & $\mathcal{D}_{f}$ & 95.30$\pm$2.08                           & 94.10$\pm$2.00 & 90.00$\pm$3.73 & 90.00$\pm$2.73                           & 59.04$\pm$4.79 & 93.61$\pm$4.99 &     94.94$\pm$2.89                    \\
    & MIA               & 75.78$\pm$0.04                           & 74.22$\pm$0.04 & 74.58$\pm$0.05 & 49.28$\pm$0.07                           & 25.18$\pm$0.05 & 72.65$\pm$0.05 &        73.86$\pm$6.33                 \\ \hline
ViT & $\mathcal{D}_{r}$ & 98.88$\pm$0.00 & 98.61$\pm$0.08                           & 97.28$\pm$0.33 & 97.58$\pm$0.36                           & 97.62$\pm$0.35 & 98.01$\pm$1.56 &          98.53$\pm$0.27               \\
    & $\mathcal{D}_{f}$ & 100.00$\pm$0.00                          & 98.80$\pm$0.76 & 97.19$\pm$0.98 & 86.75$\pm$3.57                           & 73.49$\pm$5.11 & 98.07$\pm$2.35 &     99.52$\pm$0.84                    \\
    & MIA               & 90.76$\pm$0.03                           & 91.77$\pm$0.02 & 86.14$\pm$0.02 & 33.53$\pm$0.06                           & 10.44$\pm$0.05 & 85.54$\pm$0.11 &       90.12$\pm$3.40                  \\ \hline
\end{tabular}
\end{table}

\end{document}